\documentclass[lettersize,journal]{IEEEtran}
\usepackage{amsmath,amsfonts}
\usepackage[ruled,vlined]{algorithm2e}
\usepackage{array}
\usepackage[caption=false,font=normalsize,labelfont=sf,textfont=sf]{subfig}
\usepackage{textcomp}
\usepackage{stfloats}
\usepackage{url}
\usepackage{verbatim}
\usepackage{graphicx}
\usepackage{cite, flushend}
\usepackage{subfig}        % Make sure you have this in your preamble
\usepackage[font=footnotesize]{caption} % Adjust caption font globally
\usepackage{xcolor}
\usepackage{soul}
\begin{document}

\title{Agentic AI Meets Edge Computing in Autonomous UAV Swarms}
% \title{Agentic AI and Edge Computing Unite: Toward Autonomous UAV Swarms}

\author{Thuan Minh Nguyen, Vu Tuan Truong, and Long Bao Le,~\IEEEmembership{Fellow,~IEEE}
        % <-this % stops a space
\thanks{The authors are with INRS, University of Qu\'{e}bec, Montr\'{e}al, QC H5A 1K6, Canada. Emails:  minh-thuan.nguyen@inrs.ca,
tuan.vu.truong@inrs.ca, long.le@inrs.ca.
%Manuscript received April 19, 2021; revised August 16, 2021.
}}



\maketitle

\begin{abstract}
The integration of agentic AI, powered by large language models (LLMs) with autonomous reasoning, planning, and execution, into unmanned aerial vehicle (UAV) swarms opens new operational possibilities and brings the vision of the Internet of Drones closer to reality. However, infrastructure constraints, dynamic environments, and the computational demands of multi-agent coordination limit real-world deployment in high-risk scenarios such as wildfires and disaster response. This paper investigates the integration of LLM-based agentic AI and edge computing to realize scalable and resilient autonomy in UAV swarms. We first discuss three architectures for supporting UAV swarms -  standalone, edge-enabled, and edge-cloud hybrid deployment - each optimized for varying autonomy and connectivity levels. Then, a use case for wildfire search and rescue (SAR) is designed to demonstrate the efficiency of the edge-enabled architecture, enabling high SAR coverage, reduced mission completion times, and a higher level of autonomy compared to traditional approaches. Finally, we highlight open challenges in integrating LLMs and edge computing for mission-critical UAV-swarm applications.

%efficient LLM inference, hallucination mitigation, robust multi-agent collaboration, and edge scalbi benchmarking, providing a foundation for next-generation autonomous aerial systems in  mission-critical environments.

% The integration of agentic AI, powered by large language models (LLMs) with autonomous reasoning, planning, and execution, into unmanned aerial vehicle (UAV) swarms not only unlocks new operational possibilities but also advances the vision of the Internet of Drones. However, infrastructure constraints, dynamic environments, and the computational demands of multi-agent coordination limit real-world deployment in high-risk scenarios such as wildfires. This paper examines the integration of LLM-based agentic AI with edge computing to achieve scalable, resilient UAV swarm autonomy. We discuss three deployment architectures—standalone, edge-enabled, and edge/cloud hybrid—optimized for varying autonomy and connectivity levels. A wildfire search-and-rescue (SAR) use case demonstrates the edge-enabled architecture, with simulations showing superior coverage and reduced mission completion times compared to a centralized greedy baseline. Finally, we highlight challenges in LLM inference, multimodal reasoning, decentralized collaboration, and system benchmarking, providing a foundation for next-generation autonomous aerial systems in mission-critical environments.

\end{abstract}


\section{Introduction}

% Drones, also known as unmanned aerial vehicles (UAVs), have garnered significant attention in recent years, owing to their advanced capabilities in wide-area coverage, real-time sensing, and rapid deployment.
% These capabilities render UAVs particularly effective for demanding applications, such as environmental monitoring, SAR operations, and wireless network provisioning.
% However, real-world scenarios often present dynamic conditions (e.g., changing environments) together with practical challenges, including limited GPS access, unreliable network connectivity, and restricted human supervision, especially for  drone swarms, a key technology driving the advancement of the Internet of Drones \cite{9547279}. 
% Enabling UAV swarms to operate effectively under such conditions therefore remains a significant challenge.

Unmanned aerial vehicles (UAVs), or drones, have attracted growing attention in recent years due to their capabilities in wide-area coverage, real-time sensing, and rapid deployment. When operating collaboratively as swarms, UAVs amplify these advantages and advance the vision of the Internet of Drones \cite{9547279}. Such drone swarms are particularly effective for demanding tasks, such as environmental monitoring, search and rescue (SAR) operations, and wireless network provisioning.

However, real-world UAV deployments face numerous challenges arising from dynamic conditions (e.g., environmental changes) and practical constraints such as limited Global Positioning System (GPS) access, unreliable network connectivity, and restricted human supervision. Consequently, ensuring effective UAV swarm operation under such conditions remains a major challenge. Recent advances in agentic AI, powered by large language models (LLMs) with strong reasoning capabilities \cite{NEURIPS2022_9d560961, NEURIPS2023_271db992}, offer promising solutions to these issues. {By integrating agentic AI, UAVs can autonomously reason, plan, and make decisions with minimal human intervention, enabling rapid adaptation to environmental dynamics and unforeseen events. For instance, an LLM-based agentic AI embedded within a UAV can perform functions traditionally handled by humans (e.g., navigation and control) through natural-language-driven instruction and decision-making. Moreover, a more advanced paradigm, known as multi-agentic AI \cite{kannan2024smart, feng2025ochmas}, can be employed to further enhance the capabilities of UAV swarms. In this framework, each drone functions as an autonomous agent that communicates, reasons, and collaborates with others to collectively make decisions and address complex tasks.}

{Nevertheless, recent LLMs require substantial storage, computational, and energy resources, which are often limited on small aerial systems like UAVs. In addition, real-time missions typically demand low-latency decision-making, exceeding the computational capability of UAVs' embedded hardware. Although model compression techniques such as quantization, pruning, and knowledge distillation can be employed to mitigate these challenges, they often degrade the reasoning and contextual understanding abilities of the integrated LLMs, resulting in lower mission success rates and reduced overall swarm performance.
To sustain high performance in multi-agentic AI systems without exceeding the hardware limitations of UAVs, alternative computing paradigms such as edge and cloud computing can be employed \cite{10591707, yang2025autohma}. Although cloud computing offers abundant storage and computational capacity, it often introduces considerable latency due to long-distance data transmission. Moreover, UAV swarms operating in harsh or remote environments (e.g., disaster zones or wilderness regions) may experience unstable connectivity to cloud servers, potentially causing service disruptions and mission failures. To balance the aforementioned trade-offs, edge computing offers an effective alternative by distributing inference, reasoning, and data processing across UAVs and nearby edge servers \cite{10818760}. 
However, the literature does not provide a detailed and unified account of integrating edge computing with LLM-based agentic AI for scalable, resilient, low-latency UAV swarm autonomy.
This paper aims to address this research gap by first providing background on agentic AI, followed by a description of three alternative architectures for UAV swarms—standalone, edge-enabled, and edge/cloud-enabled UAV swarms.
We then propose a hybrid, edge-enabled LLM-based agentic AI framework for the SAR use case using UAV swarms.} 

{The remainder of this paper is organized as follows. Section~\ref{sec:agentic-ai} provides background on agentic AI and presents an overview of UAV swarm design based on  practical multi-agentic AI frameworks. Section~\ref{sec:uav-swarm} discusses three deployment strategies for UAV swarms powered by multi-agentic AI, namely fully standalone, edge-enabled, and hybrid edge-cloud architectures. Section~\ref{sec:use-case} illustrates the potential of agentic-AI-powered UAV swarms through a wildfire SAR use case. Finally, Section~\ref{sec:open-challenge} highlights open challenges for such systems, and Section~\ref{sec:conclusion} concludes the paper.}

\section{Agentic AI Systems}
\label{sec:agentic-ai}

This section provides background on agentic AI with the introduction of several representative frameworks for practical implementation.

%systems, covering definitions, core components, advantages over traditional AI agents, and multi-agentic AI frameworks.

\subsection{Background on Agentic AI }

%traditional AI systems, which execute predefined tasks under rigid rules, such agentic AI embodies rationalism \cite{chong2021ai}, enabling perception, goal interpretation, decision-making, and autonomous task execution.

Agentic AI refers to intelligent systems that function as autonomous, goal-directed agents capable of reasoning, adapting, and interacting with their environment in a human-like manner.   
Agentic AI systems are particularly well-suited for complex domains like UAV swarm operations, where autonomy, adaptability, and coordinated decision-making are critical. Multi-agentic AI settings further enable collaborative behaviors among agents, allowing UAV swarms to respond collectively to evolving conditions and unforeseen challenges. In the following, we present core components of an agentic AI system.

%Additionally, agentic AI systems can communicate and collaborate with other agents and people, making them well-suited for robotics and UAV swarm applications.
%\subsection{Key Components of Agentic AI}

%Agentic AI systems are characterized by their autonomy, experiential learning, and adaptability in dynamic environments. Therefore, their core components typically include a perception layer, a reasoning module for decision making and planning, a memory module to store past experiences, a communication module to interact with humans and other agents, and an action executor for task execution.

\noindent\textbf{Perception Module.} This module captures and processes diverse inputs (e.g., natural language instructions, sensor readings, visual streams, and data from external Application Programming Interfaces (APIs)) into structured representations that can be interpreted by the reasoning module. It establishes situational awareness and contextual grounding, both of which are essential for accurate understanding and informed decision-making.

\noindent\textbf{Reasoning Module.} This module processes inputs from the perception layer to interpret objectives, decompose tasks, and generate strategic plans. Using methods such as Chain-of-Thought (CoT) reasoning and iterative self-reflection, it adapts strategies to dynamic conditions and system constraints while pursuing mission goals.

\noindent\textbf{Action Execution Module.} This module materializes plans generated by the reasoning module through interfaces with external tools, APIs, and physical actuators. It converts high-level reasoning outputs into executable actions such as code generation, API calls, and control commands for embodied agents like UAVs and robots.

\noindent\textbf{Memory Module.} This module stores past observations, interactions, and learned knowledge, which are vital for coherent multi-step reasoning and long-horizon task execution. It encompasses short-term memory, which maintains active task states and recent episodic events, and long-term memory, which accumulates experiences and contextual knowledge over time. By retaining prior interactions and outcomes, this enables experiential learning, behavioral adaptation, and continuity across sequential decision-making processes.

\noindent\textbf{Coordination Module.} This module facilitates multi-agent and human-AI interactions by managing dialogue flow, negotiating task allocations, and fostering consensus among agents, thereby enabling effective operation in distributed environments such as UAV swarms.

Modern agentic AI-based platforms such as humanoid robots and drones can leverage recent vision–language–action (VLA) models to build their perception, reasoning, and execution modules. These models are capable of translating multimodal observations and natural-language instructions directly to executable actions, such as navigation or control commands. For instance, RT-2 \cite{zitkovich2023rt} builds upon large pretrained vision–language models (VLMs) and adapts them for robotic control through fine-tuning on task-specific robotic datasets. OpenVLA \cite{kim2025openvla} extends this paradigm with an open-source, scalable architecture trained on the Open-X-Embodiment dataset, offering reproducible pipelines and adaptability across a wide range of robotic platforms.

%Let us elaborate by using concrete examples. For modern robots such as humanoid robots or drones, the perception, reasoning, and execution functions can be realized by an advanced machine learning model known as the vision–language–action (VLA) model. In particular, the VLA model maps multimodal observations and natural-language instructions directly to executable actions such as control commands for robots or UAVs. For example, the VLA model called RT-2 \cite{zitkovich2023rt} is based on large vision–language models (VLMs) pretrained on web-scale data and it is adapted for robotic control through fine-tuning using robotic data. OpenVLA \cite{kim2024openvla} extends this concept through an open-source, scalable VLA architecture trained on the Open-X-Embodiment dataset, offering reproducible training pipelines and adaptability to diverse robotic platforms. 

%More recently, $\pi_{0.5}$ \cite{intelligence2025pi_} introduces a scalable, high-performance policy that integrates perception, reasoning, and control into a compact, instruction-following model capable of rapid adaptation to varied robot morphologies and environments. These works show the potential of VLA models to unify perception, reasoning, and action within a single agentic AI system.

\subsection{Multi-agentic AI Frameworks}

A variety of LLM-based multi-agent frameworks have recently emerged to coordinate complex and collaborative workflows in robotics. In swarm robotics, these frameworks facilitate persistent state sharing, dynamic task allocation, adaptive planning, and other capabilities crucial for maintaining autonomy under real-world constraints. Several representative frameworks applicable for UAV applications are summarized below.

\noindent\textbf{LangGraph} is a graph-based, stateful, and event-driven multi-agent framework in which nodes represent agents, tools, or memory modules, and edges define control and data flows. It supports cyclic and asynchronous execution, allowing UAVs to continuously sense, plan, and coordinate without waiting for centralized directives. In swarm mode, LangGraph eliminates single points of control by enabling agents to hand off tasks directly to peers via handoff tools. This decentralized design is particularly well-suited for autonomous UAV swarms operating in unpredictable or communication-constrained environments.

%LangGraph is a graph-based, stateful, and event-driven multi-agent framework where nodes represent agents, tools, or memory modules, and edges define control/data flow. LangGraph supports cyclic and asynchronous execution, enabling UAVs to continuously sense, plan, and coordinate without halting for centralized directives. The swarm mode eliminates a single point of control, allowing agents to hand off tasks directly to peers using handoff tools. This decentralization is particularly suited for autonomous UAV swarms operating in unpredictable or communication-limited environments.

\noindent\textbf{AutoGen} is a conversation-driven framework that enables multi-agent collaboration through structured dialogue \cite{wu2024autogen}. Agents assume roles such as planner, executor, or verifier and interact via prompt engineering and message passing. In robotics, AutoGen supports high-level mission planning, tool invocation (e.g., perception or actuation APIs), and reflective reasoning via critic agents. It accommodates both human-in-the-loop (HITL) and fully autonomous operations, making it well-suited for UAV swarms. Its modular design and tool integration allow seamless interfacing with robotic subsystems, including sensor fusion and navigation modules.

%AutoGen is a conversation-driven framework developed by Microsoft that facilitates multi-agent collaboration through structured dialogue. Each agent can assume specific roles such as planner, executor, or verifier, and interact with others through prompt engineering and message passing. In robotic applications, AutoGen enables high-level mission planning, tool invocation (e.g., calling perception or actuation APIs), and reflective reasoning through critic agents. It accommodates both HITL and fully autonomous operations, making it ideal for UAV swarms with edge operator support. Its modular agent design and built-in support for external tools allow seamless interfacing with robotic subsystems, including sensor fusion modules and navigation stacks.

\noindent\textbf{CrewAI} is a lightweight, role-based framework for coordinating teams of autonomous agents. Inspired by human workflows, it assigns predefined roles (e.g., Leader, Analyzer, or Marketer) to LLM-based agents collaborating on tasks. Unlike graph-based systems, CrewAI emphasizes sequential, collaborative execution through intuitive workflows and modular role definitions. This approach is particularly well-suited for robotics applications involving modular subsystems, including perception, planning, navigation, and communication.

These frameworks differ in architectural philosophy: LangGraph uses stateful, event-driven graph orchestration; AutoGen relies on dialogue-driven coordination; and CrewAI employs role-based sequential execution. In the following, we demonstrate how LangGraph can support the practical, end-to-end design of UAV swarms, illustrating its integration with perception, reasoning, memory, and execution modules.

%CrewAI is a lightweight, role-based framework for coordinating teams of autonomous agents. Drawing inspiration from human workflows, it assigns predefined roles, such as Leader, Analyzer, or Marketer, to LLM-based agents collaborating on task execution. Unlike graph-based systems, CrewAI emphasizes sequential and collaborative task execution through intuitive workflows and modular role definitions. This structure is particularly beneficial in robotics applications that involve modular subsystems such as perception, planning, navigation, and communication. 

%These frameworks differ in architectural philosophy: LangGraph employs stateful, event-driven graph orchestration; AutoGen emphasizes dialogue-driven coordination; and CrewAI adopts role-based sequencing.
%In the following subsection, we focus on LangGraph Swarm as an end-to-end practical design example for UAV swarms, demonstrating its integration with perception, reasoning, memory, and execution modules.

\subsection{UAV Swarm Design with LangGraph }

%This section presents design guidelines and a reference implementation using LangGraph as the coordination backbone, showing how Vision-Language Models (VLMs) and vision-based object detection systems (e.g., YOLO) integrate with UAV hardware.

While the preceding section discussed multi-agent frameworks, their deployment in UAV swarms requires a systematic integration of perception, reasoning, communication, and action modules. Perception is realized by transforming multimodal UAV sensor data into structured textual representations. Onboard algorithms such as YOLO perform real-time object detection on RGB and thermal imagery, while lightweight vision-language models such as LLaVA  provide contextual descriptions. Together, these outputs enable the reasoning module to fuse symbolic and perceptual information.

In the LangGraph-based UAV swarm, perception functions operate as local graph nodes within each agent, supporting autonomous situational awareness. When necessary, processed outputs are shared via the wireless mesh network to update the persistent state, allowing peers to adjust mission plans based on shared observations.

Decision-making and task planning are distributed throughout the swarm. Each UAV includes a lightweight LLM (e.g., TinyLLaMA or Phi-3-mini) to enable low-latency plan adjustments, hazard avoidance, and local decision-making. Through LangGraph’s swarm handoff mechanism, UAVs can dynamically delegate tasks to their peers.
The action execution layer translates reasoning outputs into concrete flight and payload actions. LangGraph tool interfaces connect with the UAV platform’s native Software Development Kit (SDK), enabling simple commands to directly control navigation, hovering, landing, or payload release.

A robust memory module is essential for continuity in multistep reasoning, enabling UAVs to adapt effectively in dynamic environments. Short-term memory holds active task states and recent events for immediate decisions, while long-term memory retains mission history, environment maps, and coordination patterns to guide future task allocation and hazard response. In LangGraph, memory modules serve as persistent state stores accessible to relevant nodes, allowing agents to recall detections, mission context, and lessons from prior deployments.

A self-organizing mesh network is utilized to maintain communication and coordination within the swarm. This network allows for continuous peer-to-peer negotiation without centralized control, utilizing technologies such as Wi-Fi. The LangGraph swarm's asynchronous message-passing architecture facilitates the rapid redistribution of workloads in response to changing operational conditions, thereby enhancing system resilience and guaranteeing efficient area coverage.

%By integrating multi-agent frameworks with specific model selections, robust memory integration, sensor-processing pipelines, and hardware control strategies, this design approach offers a reproducible pathway for building operational UAV swarm systems. The resulting architecture is adaptable across robotic platforms, ensuring that advances in VLMs and embedded AI translate effectively into mission-ready deployments.

\section{Multi-Agentic AI for UAV Swarms}
\label{sec:uav-swarm}

\begin{figure*}
    \centering
    \includegraphics[width=1\linewidth]{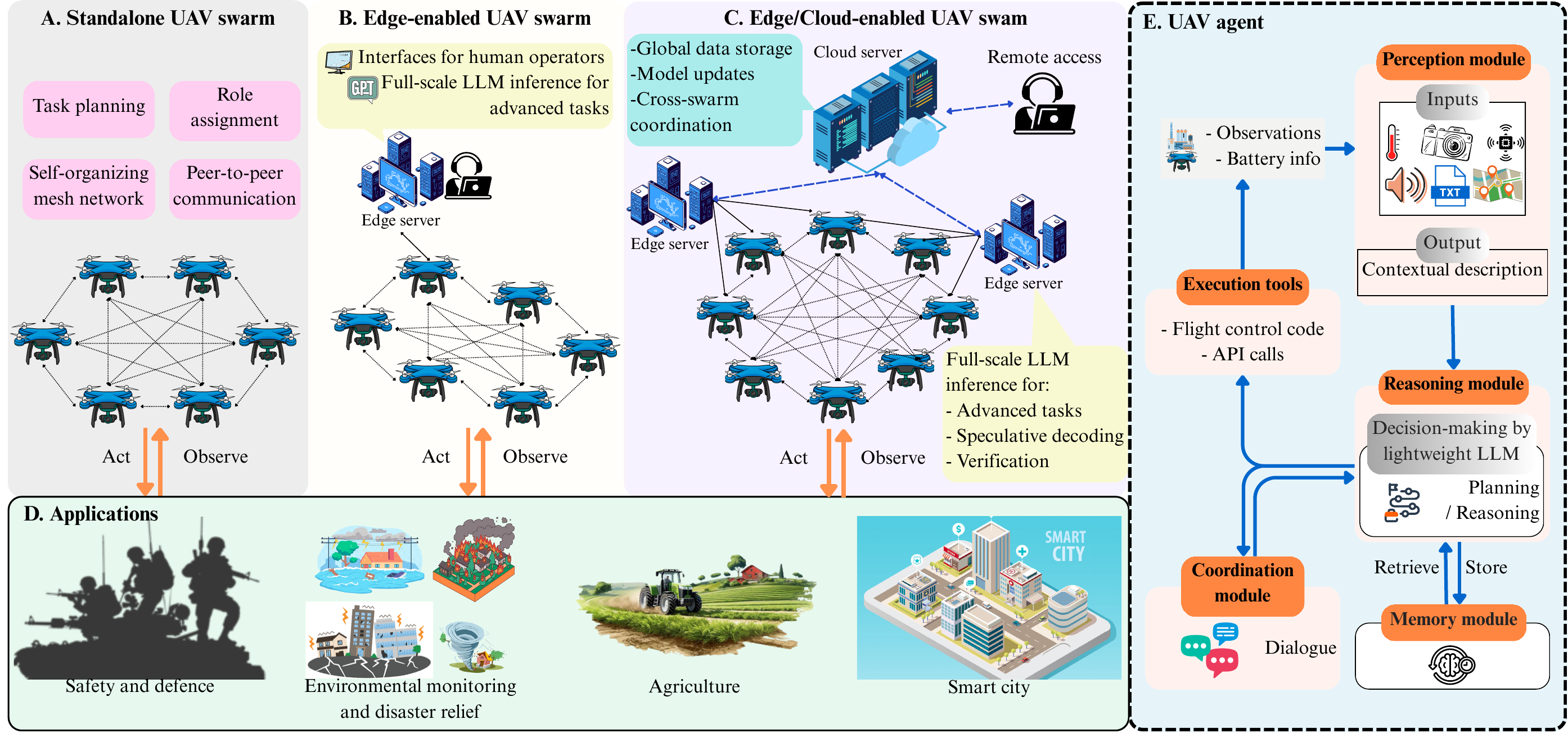}
    \caption{{Three multi-agentic AI deployment architectures for UAV swarms:
(a) Standalone UAV swarm; 
(b) Edge-enabled UAV swarm;
(c) Edge/cloud-enabled UAV swarm;
(d) Applications;
(e) UAV agent}}
    \label{fig:Agentic AI-based UAV swarm}
\end{figure*}

We now consider alternative designs for system-level deployment  of UAV swarms. Specifically, we analyze three architectures—standalone, edge-enabled, and edge/cloud-enabled UAV swarms—as illustrated in Fig. \ref{fig:Agentic AI-based UAV swarm}.

%We now consider different design alternative to system-level deployment. In particular, we analyze three deployment architectures, including standalone, edge-enabled, and edge/cloud-enabled UAV swarms, as illustrated in Fig. \ref{fig:Agentic AI-based UAV swarm}. 

%In addition, LangGraph Swarm serves as the underlying coordination framework of UAV swarms in all architectures. Each architecture offers distinct capabilities, satisfying varying levels of autonomy, collaboration, and computational demand.

%for infrastructure-denied environments with onboard reasoning and peer-to-peer coordination;
%with mobile ground station for high-level reasoning and HITL control
%integrating onboard inference, edge-level validation, and cloud-based global mission oversight

\subsection{Standalone UAV Swarm}

The first design, shown in Fig. \ref{fig:Agentic AI-based UAV swarm}(a), targets extreme environments with little or no communication infrastructure, such as disaster zones, remote wilderness, or adversarial regions.
In these scenarios, each UAV functions as a self-contained intelligent agent equipped with a multimodal sensor suite that includes optical and infrared cameras, Global Navigation Satellite System (GNSS), light detection and ranging (LiDAR), inertial measurement units (IMUs), and millimeter-wave radars for strong autonomous perception. Onboard processors such as NVIDIA Jetson Orin or Qualcomm RB5 support lightweight or quantized LLMs (e.g., TinyLLaMA), enabling onboard reasoning for decision-making and task planning.
 
The execution layer transforms reasoning outputs into specific actions by interacting with APIs and integrated actuators. Contemporary UAVs, such as the Skydio X10, already incorporate sophisticated control and navigation systems for autonomous obstacle evasion and accurate flight in GPS-denied or congested environments, allowing each UAV to sense, analyze, and perform tasks independently of external infrastructure.

 UAVs communicate through a multi-agent AI framework, enabling the swarm to function as a cohesive, goal-directed system without centralized control and to remain resilient under communication loss or unit failure. A self-organizing wireless mesh network using technology such as Wi-Fi supports mission updates, task allocation, and route synchronization with the ground station. This decentralized design enhances autonomy and adaptability but requires energy-efficient AI, lightweight embedded LLMs, and careful safeguards to mitigate risks from reduced human oversight.

%Furthermore,  UAVs of the swarm are able to communicate with each other through a  collaboration framework of multi-agentic AI, allowing the UAV swarm to operate as a cohesive, goal-directed system without reliance on centralized control, ensuring resilient operation even with intermittent communication or individual UAV failure. In particular, UAVs in a swarm establish a self-organizing mesh network using short-range wireless links (e.g., Wi-Fi Direct, LoRa) that facilitate the exchange of mission-critical updates, task assignments, planning, and routes synchronized information toward the ground station. This decentralized mechanism allows the swarm to adapt dynamically to mission objectives and environmental uncertainties. Although this approach maximizes autonomy and resilience, it demands energy-efficient AI processing and lightweight embedded LLMs and presents challenges in mission efficiency and safety due to lack of human oversight.

\subsection{Edge-enabled UAV Swarm}
\label{subsec: UAV swarms with mobile ground station support}

To enhance swarm performance while preserving decentralized autonomy, mobile ground stations serving as edge servers can be employed to support UAV swarms, as shown in
Fig. \ref{fig:Agentic AI-based UAV swarm}(b). 
In fact, mobile ground stations can maintain reliable and low-latency communication links with the UAV swarm, enabling high-level oversight and HITL collaboration. 
In addition, they provide user interfaces that enable command issuance, swarm monitoring, and real-time panoramic mapping of UAV positions and situational contexts.
Moreover, these edge servers leverage high-capacity processors to run full-scale LLMs for advanced reasoning, event validation, and mission planning, while aggregating logs and sensor data for forensic analysis, post-mission learning, and regulatory compliance. This architecture balances autonomy with flexibility, making it well-suited for semi-connected, dynamic environments.

%Moreover, these edge servers host high-capacity processors to run full-scale LLMs for advanced reasoning, event validation, and global mission planning. They also aggregate logs and sensor data for forensic analysis, post-mission learning, and compliance reporting. This architecture balances autonomy with operational flexibility, making it ideal for semi-connected, dynamic environments.

\subsection{Edge/Cloud-enabled UAV Swarm}

Fig. \ref{fig:Agentic AI-based UAV swarm}(c) illustrates the third design, intended for large-scale missions where reliable communication and integrated edge-cloud infrastructure support the UAV swarm. This design enables the UAV swarm to operate efficiently over large areas by combining onboard decision-making with powerful edge-cloud support.
In particular, each UAV is equipped with components, including multi-modal sensors and embedded lightweight LLMs for onboard reasoning, planning, and decentralized communication, as in the previous designs.

%The onboard LLM also enables the UAV to independently interpret tasks, reason, and execution in real time, and collaborate with other UAVs through swarm-level communication.

Nonetheless, lightweight onboard LLMs may experience reasoning and planning limitations—including hallucinations and reduced accuracy—which raises concerns about safety, ethics, and task reliability.
The ground stations can be leveraged to solve this problem
where UAVs run a smaller onboard LLM to produce approximate reasoning results while the edge ground station (EGS) relies on a larger LLM to validate and, if necessary, correct these outputs.
This approach enables rapid UAV decision-making, speeds up edge inference through parallel token verification, and reduces communication by sending concise outputs. Confidence-based uploads \cite{wang2023tabi} further save bandwidth, while the EGS aggregates UAV inputs to build a shared situational view and support global task redistribution.

Cloud platforms extend system capabilities with persistent data storage, global map construction, cross-swarm collaboration, and continuous model updates. The architecture also enables remote monitoring and high-level command from centralized locations. By integrating edge and cloud infrastructure, multi-agent AI-driven UAV swarms achieve scalable, intelligent, and resilient operations for large, long-duration missions.

%This method enables low-latency preliminary decision-making at the UAV level, accelerates the inference process on the edge server by using a parallel decoding process to verify long sequences of tokens generated by the UAV, and minimizes communication overhead by transmitting only concise token outputs instead of full feature sets. Additionally, confidence-based mechanisms \cite{wang2023tabi} allow UAVs to selectively upload outputs for verification, further conserving bandwidth and computational resources. Additionally, the edge server conducts advanced situational interpretation by aggregating situational descriptions from multiple UAVs, building a shared understanding of the operational environment and the operational state of UAVs, and assisting with global task redistribution.
%Trained LLM parameters are shared with edge servers to maintain consistent reasoning performance across the system.

%Cloud platforms further expand system capabilities by providing persistent data storage, global map construction, cross-swarm collaboration, and continuous model updates.  The architecture also supports remote access, allowing operators to monitor swarm status and issue high-level commands from centralized locations.
%Utilizing edge and cloud infrastructure, multi-agent AI-driven UAV swarms accomplish scalable, intelligent, and resilient operations appropriate for extensive, prolonged missions. 

%Nonetheless, guaranteeing dependable network connectivity and effective resource management is essential for optimal system performance.

\subsection{Applications}

Agentic AI–powered UAV swarms can be applied across diverse domains as follows.

\textit{Safety and Defense:} Autonomous UAV swarms enable persistent surveillance, border monitoring, and threat detection in contested or infrastructure-denied environments. Through onboard reasoning and decentralized coordination, they remain resilient and maintain situational awareness under hostile conditions.

\textit{Environmental Monitoring and Disaster Relief:} UAV swarms enable real-time environmental assessment, from air quality monitoring and wildfire tracking to search-and-rescue missions. Linked with EGSs, they rapidly identify survivors and hazard zones while adapting to dynamic disaster scenarios such as floods, earthquakes, and wildfires.

\textit{Agriculture: }  
UAV swarms support precision agriculture by monitoring crop health, analyzing soil, and managing livestock. With distributed reasoning and multi-modal sensing, they provide large-area coverage and real-time decision-making, enhancing efficiency and sustainability in farming.

\textit{Smart City:} 
In urban environments, UAV swarms support infrastructure inspection, traffic monitoring, and public safety. Cloud integration enables global coordination, data sharing, and remote access for city authorities, improving responsiveness and efficiency.

These applications highlight the versatility of multi-agent AI-enabled UAV swarms, especially when coupled with edge and cloud platforms. The architecture delivers reliable autonomy, advanced reasoning, and resilient performance across diverse real-world scenarios.

\section{Use Case: Wildfire Search-and-Rescue (SAR)}
\label{sec:use-case}
Recent wildfires, such as those in Los 
Angeles in 
January 2025\footnote{“Spread of the Palisades and Eaton Fires – January 2025,” NASA Scientific Visualization Studio, 11 July 2025. [Online]. Available: \url{https://svs.gsfc.nasa.gov/5558/}.}, highlight the limits of traditional SAR operations in dynamic, hazardous environments. To overcome these challenges, we propose an autonomous SAR system combining multi-agent AI, UAV swarms, and edge computing.

\subsection{Mission Objectives}

%\textcolor{red}

{Our proposed system is designed to enable adaptive and resilient UAV swarm operations in dynamic and hazardous wildfire environments. Specifically, the design seeks to achieve the following mission objectives.  
First, it dynamically segments wildfire zones and assigns subregions to UAVs for autonomous surveying and survivor detection.
 The design supports onboard LLM-based UAV route planning,  detect human presence and residual hazards using multimodal sensing. Also, the design allows transmitting situational awareness data from UAVs to the EGS for real-time monitoring and assessment.}

\subsection{System Architecture and Operational Workflow}

\begin{figure}
    \centering
    \includegraphics[width=1\linewidth]{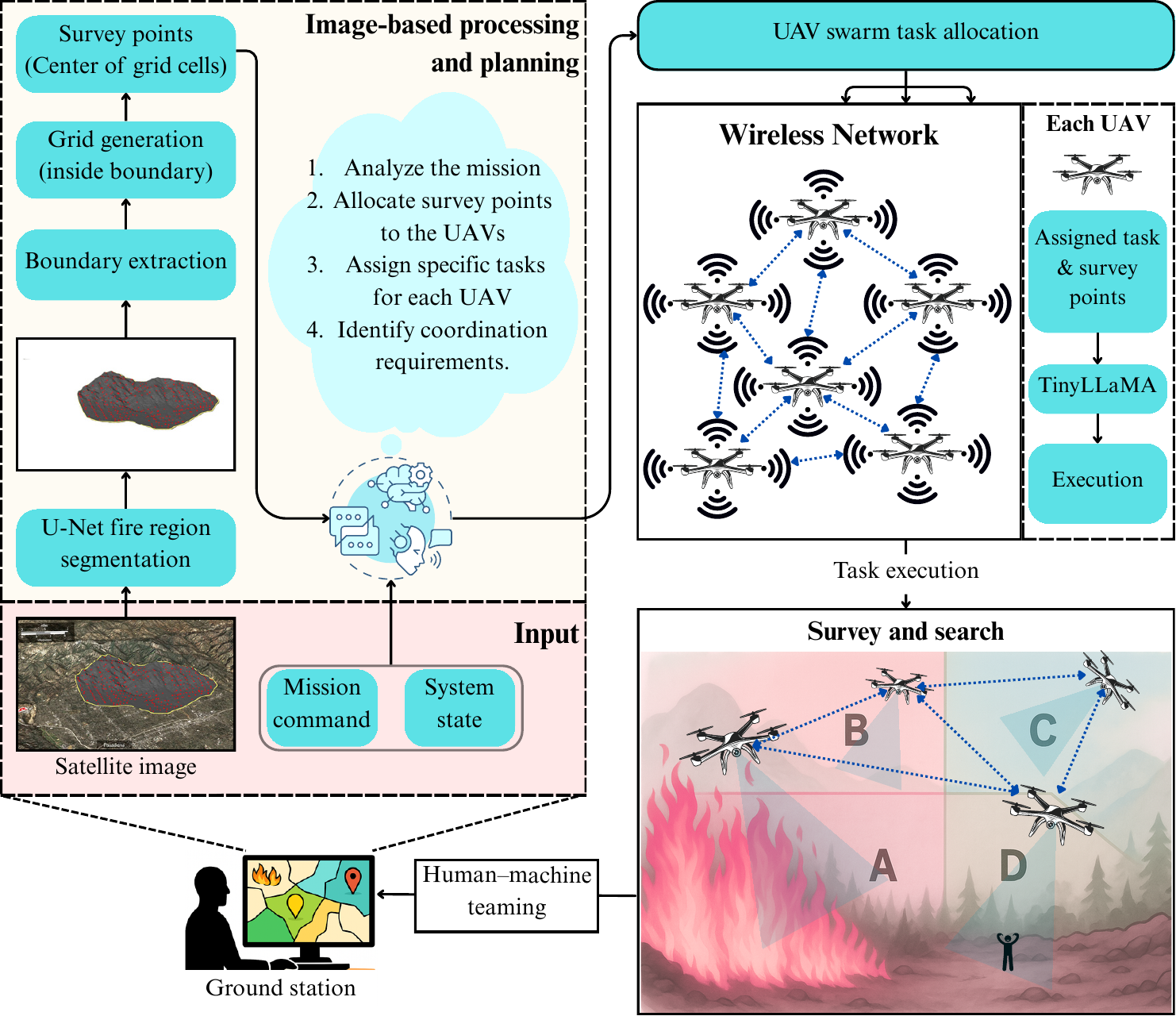}
    \caption{Proposed multi-agentic AI-driven UAV swarm system for wildfire SAR.}
    \label{fig:Operational workflow}
\end{figure}

Our proposed system for wildfire SAR integrates three core components—satellite image-based planning by an EGS, distributed swarm intelligence, and resilient human–machine teaming—into a unified workflow (Fig.~\ref{fig:Operational workflow}). It is designed for deployment on advanced UAVs such as the Skydio X10, which combines multimodal sensing, powerful onboard processing, and wireless communications. The framework is designed so that mission goals can be achieved by tackling four different tasks: 1) Image segmentation task to identify the wildfire area, 2) Survey-point creation task, 3) Survey-point assignment task, and 4) UAV path planning task. {The first three tasks are executed by the EGS to guarantee global situational awareness and effective resource coordination, whereas the final task is performed by each UAV to achieve localized autonomy and responsiveness. Detailed operations and decision-making processes of the EGS and UAVs are described in Algorithm \ref{alg:swarm}.\footnote{{To account for the UAVs’ limited energy, computing, and storage capacities, the lightweight TinyLLaMA model is used onboard for route planning, while the more capable GPT-4.1 model is employed at the EGS, which has a richer resource pool.
}
}} 

\begin{algorithm}[!ht]
\SetAlgoLined
\DontPrintSemicolon
\small
\KwIn{System state for $N$ UAVs $\mathcal{U} = \{U_1, U_2, \dots, U_N\}$, satellite image $I(t)$, ground station position $G$, mission command $M$}

\textbf{At Edge Ground  Station:}\;
\While{wildfire relief is active}{
    - Acquire latest satellite image $I(t)$\;
    - Extract fire boundary $\mathcal{B}(t)$ from $I(t)$ using U\text{-}Net segmentation model\;
    - Generate uniform grid within $\mathcal{B}(t)$ and define set of survey points $\mathcal{S}(t) = \{s_k\}$ at cell centroids\;
   -  Construct mission prompt $\mathcal{P} = (M, \mathcal{U}, \mathcal{S}(t))$\;
    % \Repeat{ $\sum_i |\mathcal{A}_i| \le |\mathcal{S}(t)|$ \textbf{and} $\sum_i |\mathcal{A}_i| \ge |\mathcal{S}(t)|$ }{
    - Obtain survey-point assignment solution via LLM-based mission planner: $\{\mathcal{A}_i\}_{i=1}^N \leftarrow \texttt{Planner}(\mathcal{P})$\;
    \uIf{$\sum_i |\mathcal{A}_i| > |\mathcal{S}(t)|$}{
        - Append ``You are hallucinating, creating more survey points than required. Do not invent, modify, or add any new points.'' to $\mathcal{P}$\;
        - Re-plan to obtain new solution of survey-point assignments via LLM-based mission planner: $\{\mathcal{A}_i\}_{i=1}^N \leftarrow \texttt{Planner}(\mathcal{P})$\;
    }
    \uElseIf{$\sum_i |\mathcal{A}_i| < |\mathcal{S}(t)|$}{
        - Append ``You have not assigned all survey points to UAVs. You must allocate all survey points to UAVs.'' to $\mathcal{P}$\;
        - Re-plan to obtain new solution of survey-point assignments via LLM-based mission planner: $\{\mathcal{A}_i\}_{i=1}^N \leftarrow \texttt{Planner}(\mathcal{P})$\;
    }
    % }

    - Transmit validated survey-point assignments $\{\mathcal{A}_i\}$ to UAVs via wireless communications links\;
    % - Transmit real-time state information from UAV swarm to the GSC\;
}

\vspace{0.25em}
\textbf{At UAVs (executed in parallel at $N$ UAVs):}\;
\For{$i = 1$ \KwTo $N$}{
    - UAV $i$ receives assigned survey points $\mathcal{A}_i$\;
    - Form local prompt $\mathcal{P}_l = (\mathcal{A}_i, \text{local perception})$\;
    % \Repeat{ $|\mathcal{T}_i| \le |\mathcal{A}_i|$ \textbf{and} $|\mathcal{T}_i| \ge |\mathcal{A}_i|$ }{
    - Generate flight route via TinyLLaMA based planner: $\mathcal{T}_i \leftarrow \texttt{TinyLLaMA}(\mathcal{P}_l)$\;
    \uIf{$|\mathcal{T}_i| > |\mathcal{A}_i|$}{
       - Append ``You have  used more survey points than required. Do not invent, modify, or add any new points.'' to $\mathcal{P}_l$\;
        - Generate new flight route via TinyLLaMA based planner: $\mathcal{T}_i \leftarrow \texttt{TinyLLaMA}(\mathcal{P}_l)$\;
    }
    \uElseIf{$|\mathcal{T}_i| < |\mathcal{A}_i|$}{
       - Append ``You have generated a flight route not including all assigned survey points. You must visit every assigned survey point.'' to $\mathcal{P}_l$\;
       - Generate new flight route via TinyLLaMA based planner: $\mathcal{T}_i \leftarrow \texttt{TinyLLaMA}(\mathcal{P}_l)$\;
    }
    % }

   - Execute $\mathcal{T}_i$: navigate waypoints, survey grids, perform real-time detection, and coordinate with peers\;
  -  Transmit state information (position, battery, progress, detections) to EGS for task updates\;
}
\caption{Workflow of Proposed Design}
\label{alg:swarm}
\end{algorithm}

{
The proposed framework operates as described below.\footnote{{Optimization objectives and constraints for the survey-point assignment and route planning optimization are integrated into the corresponding prompts for the LLMs employed by the EGS and
UAVs, respectively. Detailed prompt designs are not presented due to the space constraint.}}
Upon receiving wildfire alerts and an image of the region of interest from the satellite system, the EGS is deployed near the affected region to process incoming data and coordinate swarm operations. First, 
%to satisfy the first mission goal, 
the EGS employs the U-Net segmentation 
model\footnote{{Available at \url{https://github.com/yueureka/WildFireDetection.git}}} to delineate wildfire boundaries. As new satellite data arrives, the fire boundary maps are continuously updated, maintaining situational awareness and supporting adaptive sub-region assignment for UAVs. The delineated region is discretized into uniform grid cells whose centroids serve as survey points for swarm deployment. 
%This design ensures systematic area partitioning and prevents redundant coverage.
}

{
Subsequently, the LLM-based mission planner at the EGS analyzes segmented region and system state information such as UAVs' remaining energy and current locations to assign survey points to different UAVs.
%Specifically, the system state information  encompasses the geographic position and remaining energy of individual UAVs.
%task completion ratio, and 
%detection events of UAV.  
The prompt of the LLM-based planner is engineered to explicitly optimize the coverage efficiency, travel distance for each UAV, balance workload distribution among UAVs, and prevent overlap of UAVs' survey points.  %thus directly addressing the second and third mission goals.
%and fifth mission goals. 
To tackle the LLM hallucinations, the output of the EGS's mission planner is validated by comparing the actual number of survey points and the number of survey points assigned to all UAVs. 
When inconsistencies are detected, such as extra or missing points, the prompt is automatically updated to initiate re-planning.
{For instance, when the EGS’s LLM hallucinates by assigning more survey points than are actually available to the UAVs, we append the following statement to the prompt during re-planning: 
``You are hallucinating, creating more survey points than required. Do not invent, modify, or add any new points.'' }
 Once validated, each UAV is provided with its subregion map (i.e., assigned survey points), task instructions, and coordination requirements, establishing the foundation for autonomous swarm collaboration.
}

{
Upon deployment, the UAVs automatically establish a self-organizing wireless network, enabling resilient peer-to-peer communication and communication between UAVs and the EGS.
%, directly supporting the fifth mission goal. 
Each UAV uses the TinyLLaMA model for onboard planning, enabling it to generate and adjust its flight path based on the survey points assigned by the EGS and real-time sensory data collected onboard.
%Each UAV employs the TinyLLaMA model for onboard planning, allowing it to generate and adjust its flight route along the assigned survey points received from the EGS and real-time sensory inputs collected onboard. 
This capability enables the UAVs to achieve the mission goals described above. The flight routes generated by TinyLLaMA are verified using the same consistency checks applied to the mission planner’s output, ensuring overall mission reliability.
}

{Once the flight route is validated, the execution layer directs UAVs to their waypoints, conducts independent grid surveys, performs real-time fire and survivor detection, adapts to environmental changes, and coordinates with swarm peers as needed. Each UAV continuously transmits telemetry data, including position, battery status, task execution progress, and detection results, to the EGS. These data allow the EGS to perform real-time mission reconfiguration and ensure sustained operational efficiency.}

\subsection{Performance Metrics}

%We define key performance metrics to objectively evaluate the effectiveness of the proposed design. In our simulation, these metrics are quantitatively derived from the published specifications of the Skydio X10 where this UAV can detect people at a range of up to 1500 m.
%Based on this, we define each grid cell to be approximately 450x450 m for optimal coverage. Therefore, the total Eaton fire-affected area can be divided into nearly 300 grid cells, equivalent to 300 survey points. Key performance metrics are defined as follows:

We now define performance metrics to quantitatively evaluate the proposed design. In our simulation, these performance metrics are based on the specifications of the Skydio X10 drone, which allows human detection up to 1500 m. Accordingly, each grid cell is set to approximately 450 m × 450 m for optimal coverage, dividing the Eaton fire-affected area into about 300 grid cells, corresponding to 300 survey points. The performance metrics are defined as follows:
\begin{itemize}
    \item \textbf{Coverage Rate:} Ratio of the number of survey points visited to the total number of survey points.
    \item \textbf{Mission Completion Time:}
{Total time elapsed from assigning tasks to the swarm until all SAR objectives—i.e., all grid cells or survey points have been visited by the UAVs—are completed.}
    
   % Total elapsed time from task allocation to the swarm until the completion of all assigned SAR objectives (e.g., all grid cells/survey points have been visited).
    % \begin{equation}
    %     Mission\_completion\_time = t_{end}-t_{start}
    % \end{equation}
    % \item \textbf{Power Consumption: } Total electrical energy expended by the UAV swarm during the mission is calculated as follows:
    % \begin{equation}
    %     E_{Consumed}(Wh)=\sum_{i=1}^{N}(E_{start,i}-E_{end,i}+k*E_{full,i}),
    % \end{equation}
    % where \(E_{start,i}\), \(E_{end,i}\), and \(E_{full}\) are the initial, final, and full battery energy readings, respectively, for the UAV \(i\), and \(k\) is the number of trips returning to base for charging. 
    
    % In our simulation, we use the manufacturer’s battery specs and measured flight durations and distances. In addition, we assume that TinyLLaMa model's idle power consumption is 5 Wh, and TinyLLaMa model's inference-conducting power consumption is 15 Wh.

\end{itemize}

%Beyond these metrics used to preliminarily evaluate our proposed system in simulation, there are other metrics, such as:
    
%\begin{itemize}
 %   \item \textbf{Detection accuracy:} Proportion of correctly detected targets (e.g., survivors) to total targets present.
%    \item \textbf{Power Consumption: } Total electrical energy expended by the UAV swarm during the mission.
 %   \item \textbf{Swarm Coordination Efficiency:} Evaluated by measuring redundant coverage, workload balance, and total inter-UAV communication. These evaluation metrics include task overlap ratio, average communication volume, and workload variance.
%\end{itemize}

\subsection{State-of-the-Art Baseline} 

{
 We adopt the baseline  \cite{chen2024scalable} for performance comparison with our design. In this work,
 the authors proposed an efficient LLM hybrid planning framework, which combines centralized planning with distributed feedback, achieving a high task success rate and desired scalability in various warehouse scenarios.
 Their performance evaluation showed that the hybrid LLM planning framework outperforms other centralized designs. 
 %We will choose this framework as a baseline and 
 We compare this baseline with our proposed design using the two performance metrics described above.
}

 %conducted a comprehensive comparison of decentralized, centralized, and hybrid multi-agent system frameworks in terms of task success rate and collaborative efficiency. The study demonstrated that the hybrid framework, which combines centralized planning with distributed feedback, achieved the highest task success rate and optimal scalability across various warehouse scenarios, particularly as the number of robots increased. 

\subsection{Simulation Setting}

{
Experiments were conducted on an Intel Core i7-10750H CPU (2.60 GHz), 16 GB RAM, and an NVIDIA GeForce GTX 1650 Ti GPU running Windows 11 (64-bit). The software environment used Python 3.11.13 and PyTorch 2.7.1 with CUDA 12.8. To simulate the full-scale LLM at the EGS, we employ GPT-4.1 via the OpenAI API for survey-point assignments, ensuring that survey points assigned for different UAVs are non-overlapping and optimized for route planning efficiency. Each UAV used TinyLLaMA-1.1B (int4) for onboard route planning, enabling it to minimize flight distance over its assigned survey points.
 We assume that each UAV simulates a Skydio X10-class platform with RGB + thermal sensors (1500 m range), a cruise speed of 15 m/s, and a 9600 mAh battery. The onboard processor modeled a Jetson Orin NX (70 TOPS, 10–25 W), with base power 45 W, flight power 8 W per 1 m/s, TinyLLaMA idle power 5 W, and inference power 10 W. Satellite images were obtained from the sped-up NASA video footage. We assume that consecutive satellite images are taken from the video, one image for every interval of 0.37 seconds. {These extracted images are segmented by the EGS to identify wildfire boundaries, enabling survey points to be obtained dynamically.}
 Note that this interval corresponds to tens of minutes in reality due to the sped-up nature of the video clip.}

\subsection{Experimental Results}

% Experiments were conducted on an Intel Core i7-10750H CPU (2.60 GHz), 16 GB RAM, and an NVIDIA GeForce GTX 1650 Ti GPU, running Windows 11 (64-bit). The software environment includes Python 3.11.13 and PyTorch 2.7.1 with CUDA 12.8 support. To simulate the full-scale LLM at the EGS, we employ GPT-4.1 via the OpenAI API, where a system prompt is  designed to enable efficient assignment of survey points to UAVs, meeting the following requirements: 1) assigned points for UAVs are non-overlapping, and 2) the assignment solution allows efficient route planning for UAVs.
% We also design the system prompt for the onboard TinyLLaMA model, instructing it to generate a UAV route that covers all assigned survey points while minimizing total travel distance.
% Satellite images were obtained from the sped-up NASA video footage. We assume that consecutive satellite images are taken from the video, one image for every interval of 0.37 seconds. Note that this interval corresponds to tens of minutes in reality due to the sped-up nature of the video clip.

%Our experiments are conducted on a Dell Inspiron 7501 equipped with an Intel Core i7-10750H CPU (2.60 GHz), 16 GB RAM, and an NVIDIA GeForce GTX 1650 Ti GPU, running Windows 11 (64-bit). The software environment includes Python 3.11.13 and PyTorch 2.7.1 with CUDA 12.8 support. Beyond that, in order to simulate the full-scale LLM at the ground station, we use model GPT-4.1 through the OpenAI API. In addition, satellite images are collected from video published by NASA \cite{NASA_SVS_5558}.

\begin{figure}[!ht]
    \centering
    % First row
    \subfloat[Initial deployment phase, with all 10 UAVs stationed at the ground control station and the survey-point assignments generated by the EGS, ready for launch.]{%
        \includegraphics[width=0.48\textwidth]{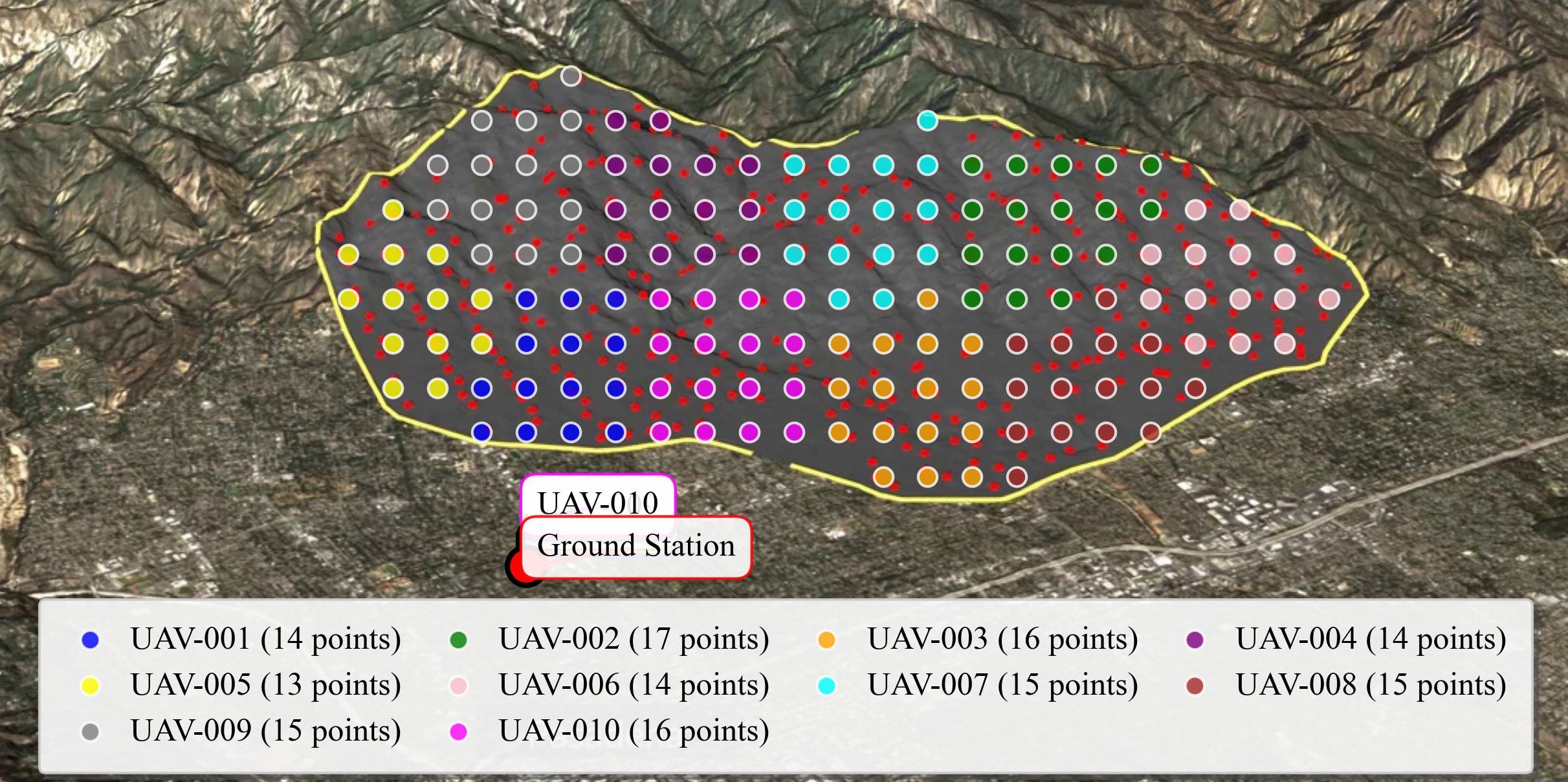}%
        \label{fig:a}
    }
    \hfil
    \subfloat[Operation phase, with UAVs dispersed and flying along their planned routes.]{%
        \includegraphics[width=0.48\textwidth]{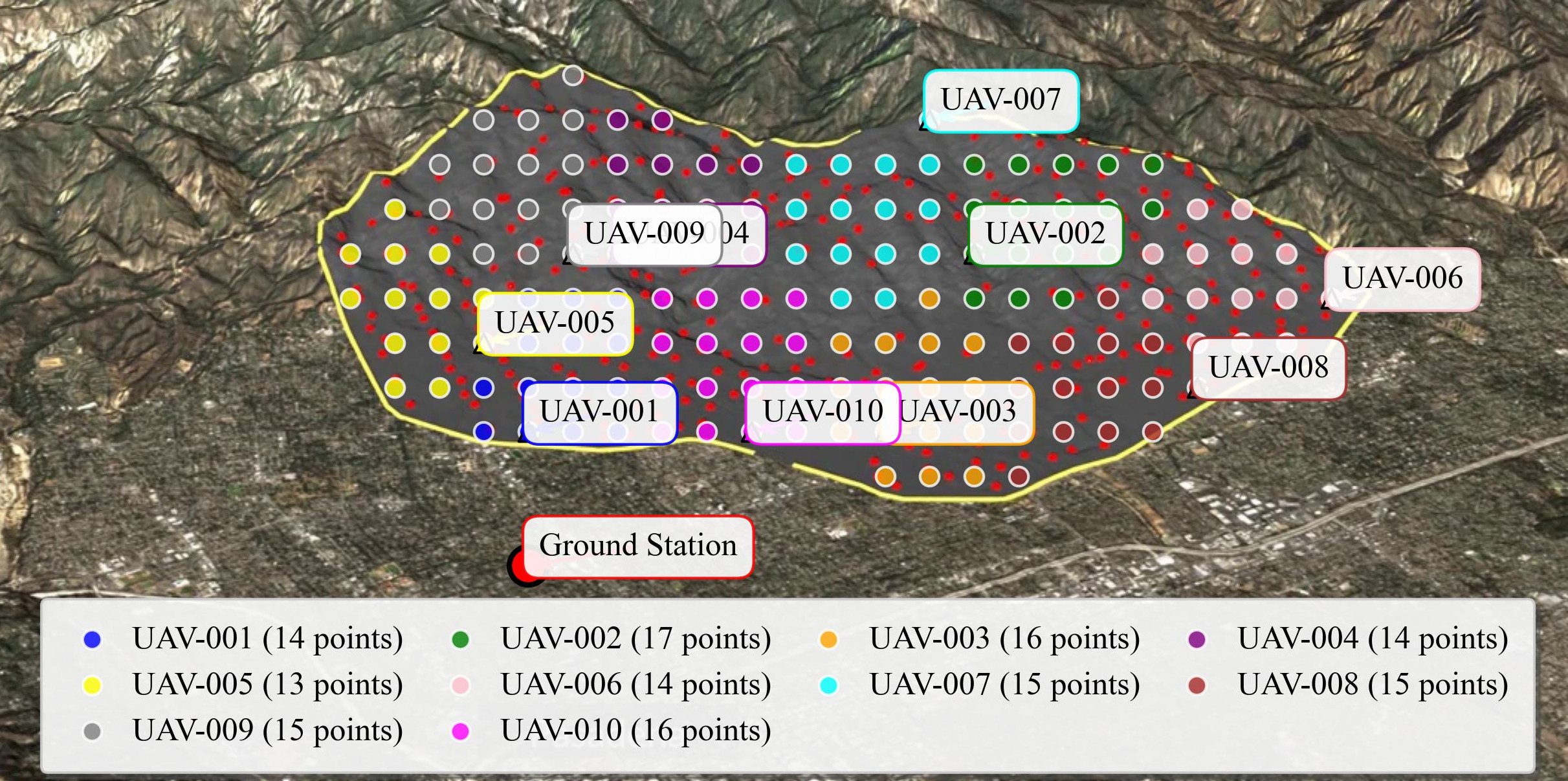}%
        \label{fig:b}
    }
    % \par\medskip % Space between rows
    \hfil
    % Second row
    \subfloat[
    Fire spread requiring dynamic fire boundary calculation and updated survey-point assignments for UAVs.]{%
        \includegraphics[width=0.48\textwidth]{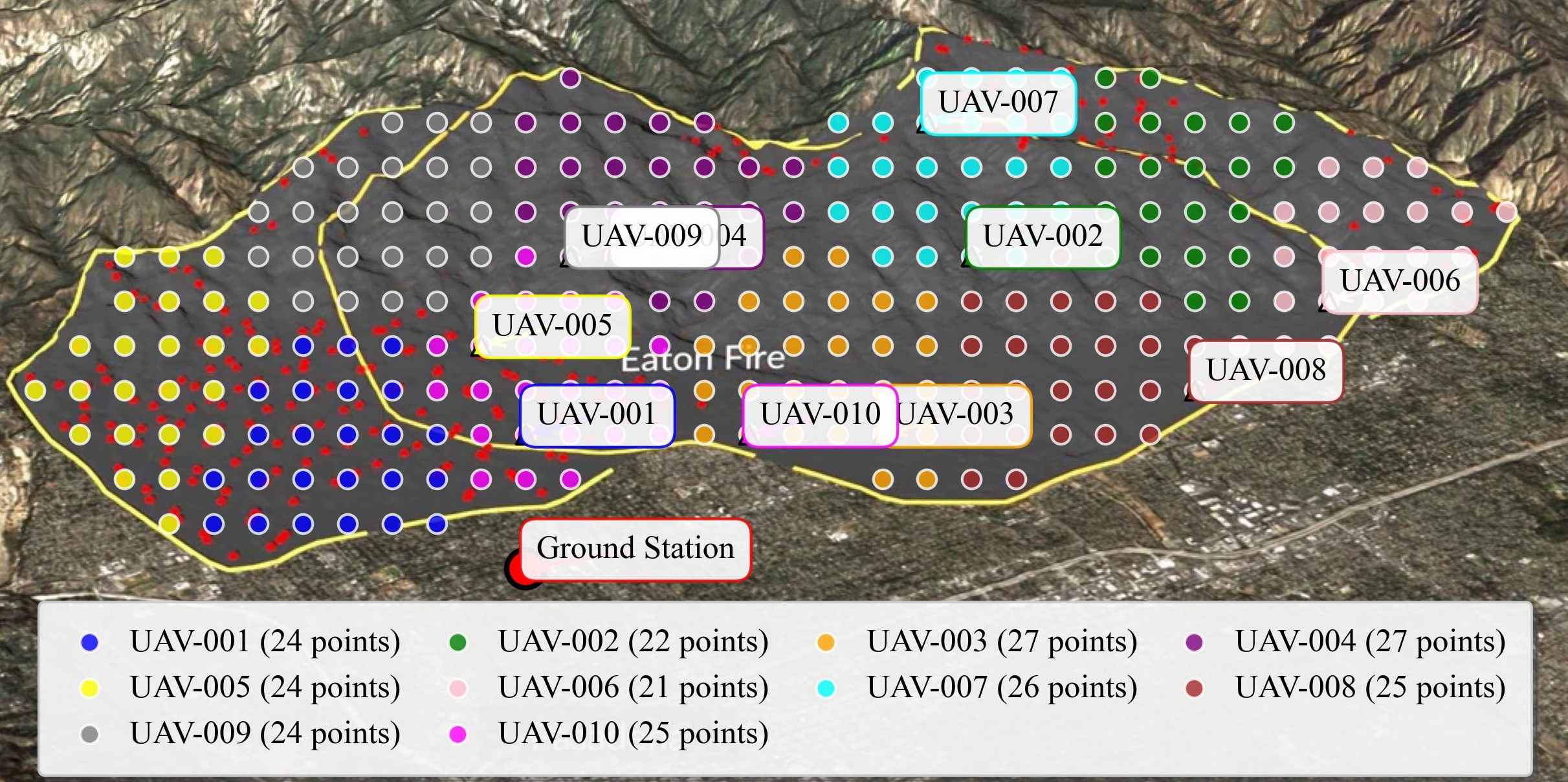}%
        \label{fig:c}
    }
    \caption{Visualization of survey-point assignments for 10 UAVs across three different phases, shown alongside the expanding wildfire region.}
%(a) Initial deployment phase with all UAVs stationed at the ground station;
%(b) Mid-operation phase with UAVs dispersed and executing assigned tasks;
%(c) Large-scale fire spread requiring dynamic reallocation and additional swarm coordination.}
    \label{fig:2x2images}
\end{figure}

%We first assess the efficiency of wildfire area segmentation and the performance of GPT-4.1 for UAV sub-region allocation across three operational milestones: (1) initial deployment, where all UAVs are stationed at the ground station; (2) mid-operation, with UAVs dispersed across affected areas; and (3) large-scale fire spread, requiring dynamic reallocation. As shown in fig. \ref{fig:2x2images}, our proposed system can effectively segment the wildfire area and allocate subregions to UAVs, ensuring operational efficiency, avoiding overlap, balancing workload, and adapting effectively to changes in environmental conditions and UAV swarm status.

We demonstrate the wildfire segmentation and GPT-4.1–based survey-point assignment solutions across three different phases: (1) initial phase with all 10 UAVs still stationed at the ground station and survey-point assignment solution generated by the EGS, ready for launch, as shown in Fig.~\ref{fig:2x2images}(a), (2) operation phase with UAVs dispersed and flying along their planned routes, as shown in Fig.~\ref{fig:2x2images}(b), and (3) active fire spread requiring dynamic fire boundary calculation and updated survey-point assignments for UAVs, as shown in Fig.~\ref{fig:2x2images}(c). As demonstrated in Fig.~\ref{fig:2x2images}, the proposed framework effectively segments the wildfire region, assigns survey points in clusters to UAVs, and ensures efficiency, minimal overlap, balanced workload, and adaptive response to wildfire changing conditions and swarm status.

\begin{figure}[!ht]
    \centering
    \includegraphics[width=1\linewidth]{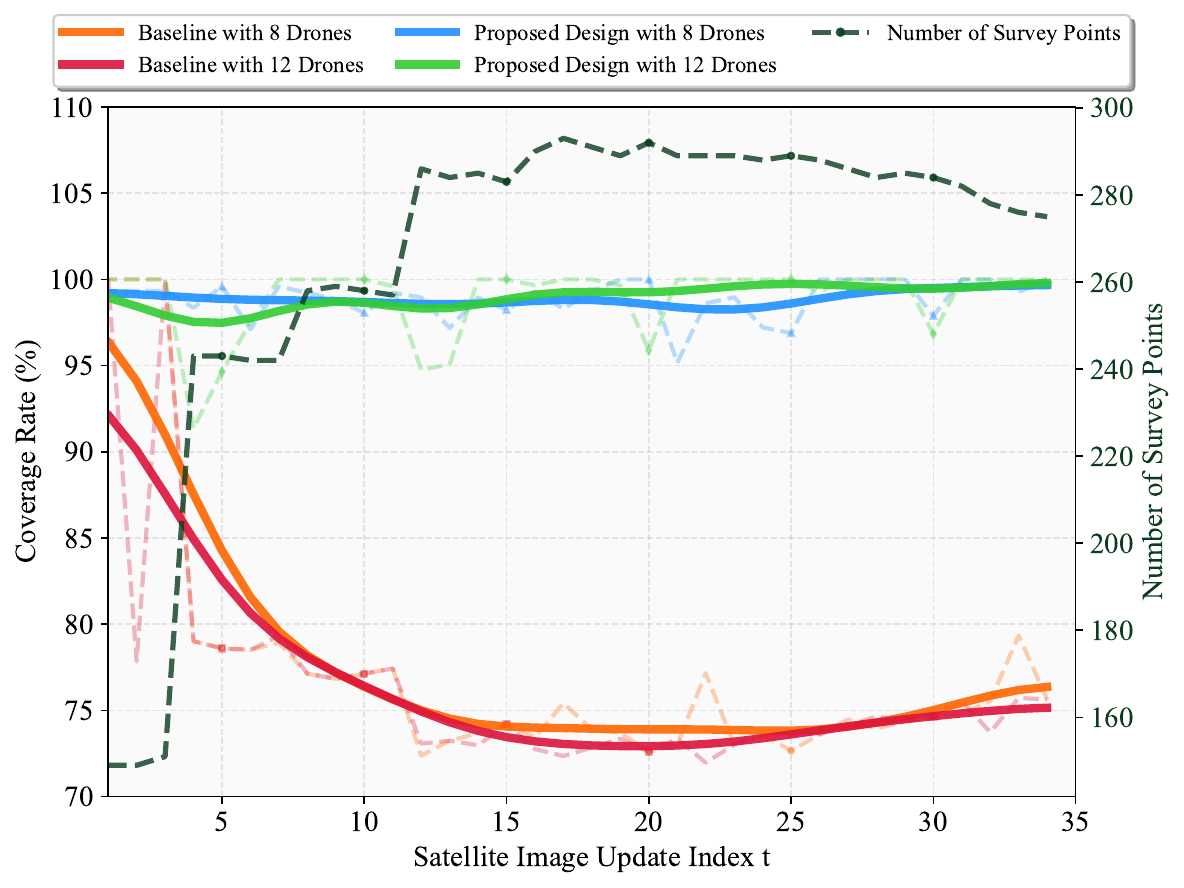}
    \caption{Coverage rate of proposed design and the baseline [13]}
    \label{fig:coverage_rate}
\end{figure}

We analyze the coverage rate achieved by the proposed design with different UAV fleet sizes (8 or 12 drones) to assess its scalability, adaptability to dynamic wildfire conditions captured at different satellite image update index $t$ (i.e.,
different time points during the wildfire period). 
As shown in Fig.~\ref{fig:coverage_rate}, the number of survey points varies over the satellite image update index $t$, reflecting the dynamic evolution of the wildfire region. 
The coverage rate achieved by our design remains high across the satellite image update indices, despite the increasing number of survey points caused by the expanding wildfire region.
 For moderate wildfire regions in small $t$, fewer UAVs (e.g., 8 UAVs) are preferable, as GPT-4.1 allocates survey points to UAVs more efficiently. Over-deployment (e.g., 12 UAVs) can cause hallucinations, generating false survey points, missing survey points, or creating overlaps. Conversely, larger wildfire regions require more UAVs (e.g., 12) to distribute the workload effectively, as too few UAVs overload the onboard TinyLLaMA reasoning model, resulting in inefficient path planning and missed survey points.
Therefore, selecting an appropriate fleet size relative to the wildfire scale is essential for efficient task allocation and mission planning.
{
It can be observed that our proposed design significantly outperforms the baseline \cite{chen2024scalable} in terms of coverage rate. This improvement is attributed to the efficient coordination between the UAVs and the EGS, as well as the validation, feedback, and adaptive re-planning strategies employed in our framework to mitigate LLM hallucination.}

%Next, we analyze the system’s coverage rate under varying UAV fleet sizes to assess scalability, adaptability to dynamic wildfire conditions, and task allocation challenges. As shown in Fig. \ref{fig:coverage_rate}, the coverage rate remains consistently high across different satellite update intervals, despite the increasing number of survey points as the fire perimeter expands. For smaller affected areas, deploying fewer UAVs (e.g., 8 UAVs) is more effective, as GPT-4.1 can allocate tasks with greater precision. Over-deploying UAVs in such cases can lead to hallucinations, where GPT 4.1 generates fake waypoints, misses real survey points, or assigns overlapping tasks. Conversely, as the wildfire expands, a larger fleet (e.g., 12 UAVs) becomes necessary to distribute the workload effectively. Insufficient UAVs in large-scale scenarios overload the local TinyLLaMA model, resulting in inefficient path planning and missing survey points. Therefore, it is crucial to dynamically adjust the number of deployed UAVs according to the scale of the wildfire to sustain both task allocation accuracy and overall mission effectiveness.

%We compare the mission completion time of our proposed design with that due to a greedy algorithm described in Algorithm \ref{alg: Greedy Algorithm}. This greedy algorithm 
%iteratively assigns survey points to UAVs with shortest distance while balancing the workload among UAVs with the set of survey points assigned to each UAV $\mathcal{A}_i, i = 1, 2, ..., N$ set empty.
%The assignment of point $j$ to UAV $i$ is based on an assignment metric $C_{ij}$, which is the weighted sum of distance and workload penalty.

To demonstrate the efficiency of our LLM-based strategy for assigning survey points to UAVs, we compare the mission completion time of our proposed design with that of a greedy assignment strategy as well as the baseline in \cite{chen2024scalable}.
%(Algorithm~\ref{alg: Greedy Algorithm}),
This greedy strategy iteratively assigns survey points to UAVs based on an assignment metric, which is a weighted sum of the (UAV, survey point) distance and a UAV's workload penalty. We define each UAV’s  set of assigned survey points as $\mathcal{A}_i$ ($i = 1, 2, \dots, N$), which is initially empty. For each assignment, the survey point $P_j$ is assigned to UAV $U_i$ if the assignment metric $C_{ij}$ is the smallest among all available (UAV, survey point) pairs.
%, defined as a weighted sum of distance and workload penalty.
Let $\bar{n} = \frac{1}{N} \sum_{i=1}^{N} |\mathcal{A}_i|$ be the average number of  survey points assigned per UAV; then the
workload penalty is defined as
\begin{equation}
 \text{Penalty}(U_i) = \max(0, |\mathcal{A}_i| - \bar{n}) \cdot B,   
\end{equation}
where $B$ is a coefficient set as $B = 800$ in the 
simulation, and {$|\mathcal{A}_i|$ represents the cardinality of set $\mathcal{A}_i $.}
The assignment metric is defined as
\begin{align}
C_{ij} = & \:  \text{Distance}(U_i.\text{current\_pos}, P_j) 
             + \lambda \cdot \text{Penalty}(U_i),
\end{align}
where $\text{Distance}(U_i.\text{current\_pos}, P_j)$ denotes the distance between the current position of UAV $U_i$ and survey point $P_j$, and $\lambda$ is the weighting parameter.

\begin{figure}[!ht]
    \centering
    \includegraphics[width=1\linewidth]{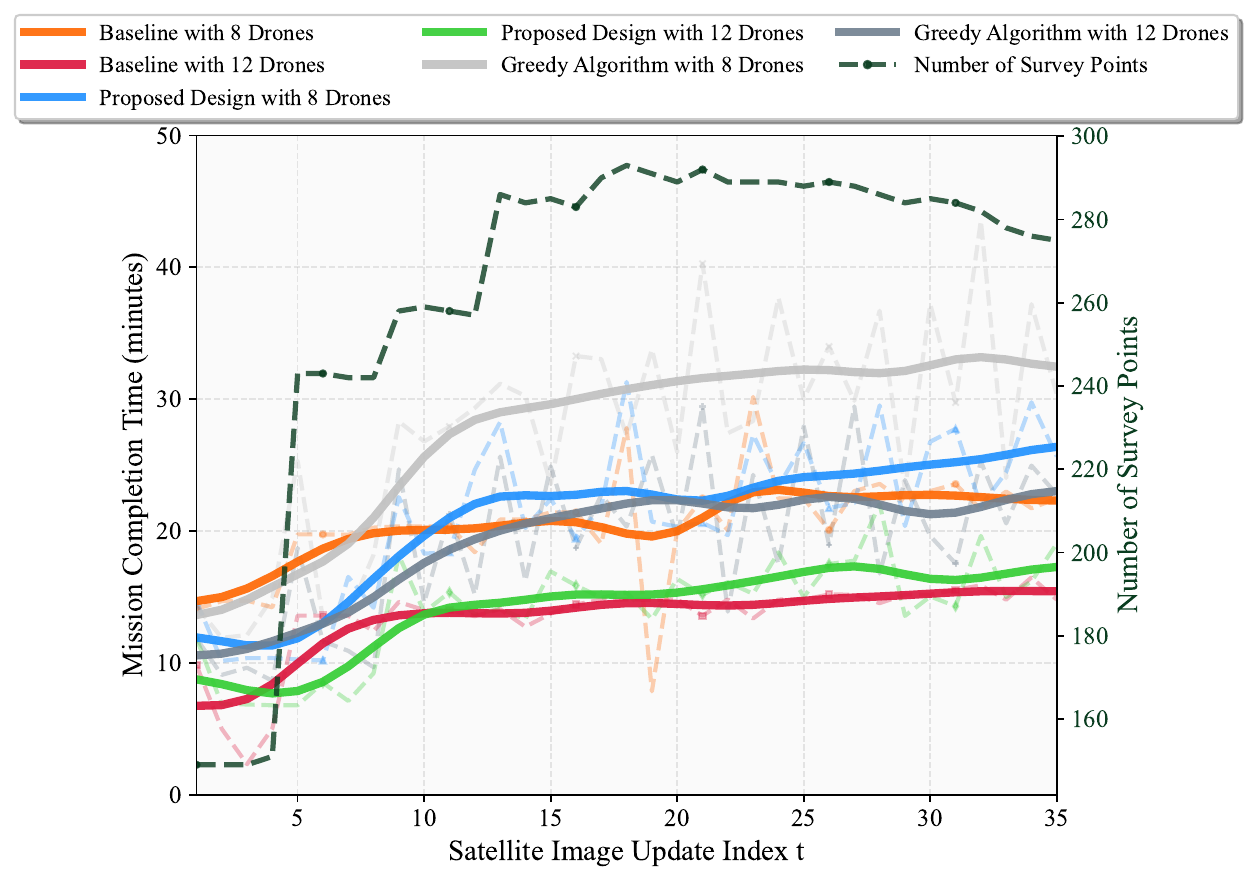}
    \caption{Mission completion time of proposed design and other baselines}
    \label{fig:Mission_time}
\end{figure}

As shown in Fig.~\ref{fig:Mission_time}, our design consistently achieves shorter mission completion time across all satellite image update indices $t$ compared to that due to the greedy strategy, especially as the number of survey points increases.  {Moreover, our proposed framework and the baseline in \cite{chen2024scalable} achieve similar mission completion time}.
With 8 UAVs, missions finish in under 25 minutes for our design, compared to over 30 minutes for the greedy algorithm. Scaling up to 12 UAVs reduces the completion time to under 17 minutes, compared with over 22 minutes for the greedy baseline.
 These results demonstrate the scalability and efficiency of our approach in dynamic, real-world conditions.

%As shown in Fig. \ref{fig:Mission_time}, our system consistently achieves shorter completion times across all wildfire milestones, particularly as the number of survey points increases. With 8 UAVs, the mission completes in under 25 minutes, while the Greedy Algorithm exceeds 30 minutes as the fire spreads. When scaled to 12 UAVs, our system reduces completion time to under 17 minutes, compared to over 22 minutes with the Greedy baseline. These results highlight the scalability and efficiency of our approach in dynamic, real-world conditions.

\section{Open Challenges}
\label{sec:open-challenge}

%Combining LLM-based multi-agent AI and edge computing is a big leap forward for autonomous UAV swarms in disaster response, 

%We discuss some open challenges for further research in the following.

We outline key open challenges for future research in the following.

\subsection{Efficient Onboard LLMs}
Although lightweight LLMs like TinyLLaMA are emerging, achieving reliable, low-latency reasoning on resource-constrained UAVs remains challenging. Future research could explore efficient model design and inference techniques, such as quantization, knowledge distillation, sparsification, and structural optimization, to enable practical onboard reasoning. Additionally, hardware–software co-design offers a promising direction for robust onboard autonomy.

\subsection{Hallucination Mitigation}

Hallucination remains a critical challenge in LLMs and LLM-driven agentic AI systems, referring to the generation of factually inaccurate content \cite{farquhar2024detecting}. In multi-agent systems, this problem is amplified through inter-agent communication: a hallucination by one agent can propagate across the network, triggering cascades of misinformation. Mitigation requires both detecting and correcting errors at the individual agent level and controlling information flow to prevent systemic error amplification. Thus, addressing hallucination in multi-agent AI demands a combination of local robustness and global coordination.

%Hallucination remains a critical challenge in LLMs and LLM-driven single-agent systems, referring to the generation of factually inaccurate content \cite{huang2025survey}. In multi-agent configurations, this issue becomes more complex due to inter-agent communication. A hallucination by one agent can propagate through the network, triggering a cascade of misinformation. Mitigating such an issue requires not only detecting and correcting errors at the individual agent level but also managing information flow across agents to prevent systemic error amplification. Addressing hallucination in multi-agentic AI thus demands both local robustness and global coordination.

% \subsection{Multi-Agent Collaboration in Harsh Environments}
\subsection{Robust Multi-Agent Collaboration}

Real-world disaster zones often feature harsh conditions, such as damaged communication infrastructure, GPS failures, and unpredictable environments. Ensuring resilient UAV swarm coordination, self-organizing mesh networks, and reliable peer-to-peer communication under these circumstances requires further research in decentralized reasoning, self-healing networks, and fault-tolerant protocols.

% \subsection{Multimodal LLM-based Multi-agentic AI}

% Most prior research on LLM-based multi-agent AI has focused on text environments, demonstrating strong language reasoning \cite{guo2024large}. However, in autonomous UAV swarms, agentic AI faces challenges in processing real-time multi-modal data, including LiDAR, RGB, and thermal imagery. Future research should address energy-efficient, low-latency multimodal fusion pipelines compatible with LLMs or the design of lightweight, efficient multimodal LLMs.

\subsection{Edge Infrastructure Scalability and Reliability}

Deploying mobile edge ground stations that support full-scale LLM inference, situational visualization, and HITL integration—while remaining portable and resilient in extreme conditions—remains a technical challenge. Future designs must balance computational power, energy efficiency, and operational resilience.

\subsection{Evaluation and Benchmarks}

Despite growing interest in LLM-based multi-agent AI, most research focuses on evaluating individual agents within narrow contexts \cite{xu2024magic}. Comprehensive benchmarks for LLM-based UAV swarm autonomy are lacking, with no established metrics for reasoning accuracy, collaboration efficiency, or mission success in realistic SAR scenarios. Developing simulation platforms, conducting field tests, and designing evaluation frameworks are essential for a thorough performance assessment.

{
\subsection{UAV Technical Limitations}
Despite advances in autonomy and swarm coordination, current UAV platforms face several hardware and operational constraints that impact mission performance and scalability. Flight time and energy limitations restrict the coverage area and duration of missions, particularly in dynamic environments. Payload and sensor constraints introduce trade-offs between weight, sensing capabilities, and mission requirements. Communication and computation bottlenecks can reduce responsiveness and efficiency, especially when relying on lightweight edge devices. These constraints affect the planning, execution, and coordination of UAV swarms, influencing mission completion time and reliability. Addressing these limitations through energy-aware planning, adaptive payload management, and hierarchical computation offloading remains  critical challenges for practical deployments.}

%Despite the increasing interest in LLM-based multi-agent AI systems, much of the current research concentrates on assessing the comprehension and reasoning of individual agents within narrowly specified contexts \cite{xu2023magic}. A significant deficiency exists in the establishment of comprehensive benchmarks for LLM-based multi-agent UAV swarm autonomy; there is an absence of metrics for reasoning accuracy, collaboration efficiency, and mission success in realistic SAR situations. Creating simulation platforms, field test methods, and evaluation frameworks is crucial for a thorough assessment of system performance.

\section{Conclusion}
\label{sec:conclusion}

{This paper investigated the integration of agentic AI and edge computing to realize scalable and resilient autonomy in UAV swarms. We discussed three swarm deployment strategies: standalone, edge-enabled, and edge/cloud hybrid UAV swarms. A wildfire SAR use case was studied to demonstrate the efficiency of the edge-enabled architecture, enabling high SAR coverage, reduced mission completion time, and a higher level of autonomy compared to existing approaches.}

{The results demonstrate that the edge-enabled architecture effectively adapts to dynamic environments while balancing autonomy and resource efficiency. 
%This study offers three mission-oriented agentic AI architectures for UAV swarms, highlights the crucial role of edge computing in overcoming UAV resource limitations and complex dynamic conditions, and
Moreover, we identified essential research challenges to guide future investigations. These findings establish a technical foundation for advancing next-generation intelligent swarm systems and provide guidance for researchers and stakeholders in robotics, edge AI, and disaster response seeking to deploy UAV swarms in real-world missions.}

%This article has investigated the integration of multi-agentic AI and edge computing for enhancing UAV swarm autonomy. We summarized the foundations of agentic AI systems and proposed three deployment architectures, along with practical applications in defence, disaster response, agriculture, and smart cities. We have also presented a specific use case, which is wildfire search and rescue, to showcase the potential of multi-agentic AI-based UAV swarms with EGS support for disaster relief operations. Finally, we discussed open challenges, including efficient LLM inference, robust swarm collaboration, and scalable infrastructure. We hope this work offers recommendations for advancing intelligent, autonomous UAV swarms for real-world missions.

% \section*{Acknowledgments}

% {\appendix[Proof of the Zonklar Equations]
% Use $\backslash${\tt{appendix}} if you have a single appendix:
% Do not use $\backslash${\tt{section}} anymore after $\backslash${\tt{appendix}}, only $\backslash${\tt{section*}}.
% If you have multiple appendixes use $\backslash${\tt{appendices}} then use $\backslash${\tt{section}} to start each appendix.
% You must declare a $\backslash${\tt{section}} before using any $\backslash${\tt{subsection}} or using $\backslash${\tt{label}} ($\backslash${\tt{appendices}} by itself
%  starts a section numbered zero.)}

\bibliographystyle{IEEEtran}
% \nocite{*}
\bibliography{references}

@inproceedings{xu2024magic,
  author    = {Xu, Lin and Hu, Zhiyuan and Zhou, Daquan and Ren, Hongyu and Dong, Zhen and Keutzer, Kurt and Ng, See-Kiong and Feng, Jiashi},
  title     = {{MAgIC}: Investigation of large language model powered multi-agent in cognition, adaptability, rationality and collaboration},
  booktitle = {Proc. Conf. Empir. Methods Nat. Lang. Process. (EMNLP)},
  month     = {Nov},
  year      = {2024},
  pages     = {7315--7332}
}

@inproceedings{wang2023tabi,
  author    = {Wang, Yiding and Chen, Kai and Tan, Haisheng and Guo, Kun},
  title     = {Tabi: An Efficient Multi-Level Inference System for Large Language Models},
  booktitle = {Proc. Eur. Conf. Comput. Syst. (EuroSys)},
  month     = {May},
  pages     = {233--248},
  year      = {2023},
  doi       = {10.1145/3552326.3587438}
}

@inproceedings{wu2024autogen,
  title={{AutoGen}: Enabling next-gen {LLM} applications via multi-agent conversations},
  author={Wu, Qingyun and Bansal, Gagan and Zhang, Jieyu and Wu, Yiran and Li, Beibin and Zhu, Erkang and Jiang, Li and Zhang, Xiaoyun and Zhang, Shaokun and Liu, Jiale and others},
  booktitle={Proc. Conf. Lang. Model. (COLM)},
  year={2024},
  month = {Oct}, 
}

@inproceedings{kim2025openvla,
  author    = {Kim, Moo Jin and Pertsch, Karl and Karamcheti, Siddharth and Xiao, Ted and Balakrishna, Ashwin and Nair, Suraj and Rafailov, Rafael and Foster, Ethan P. and Sanketi, Pannag R. and Vuong, Quan and Kollar, Thomas and Burchfiel, Benjamin and Tedrake, Russ and Sadigh, Dorsa and Levine, Sergey and Liang, Percy and Finn, Chelsea},
  title     = {{OpenVLA}: An Open-Source Vision-Language-Action Model},
  booktitle = {Proc. Conf. Robot Learn. (CoRL)},
  volume    = {270},
  pages     = {2679--2713},
  year      = {2025},
  month = 	 {06--09 Nov}
}

@inproceedings{zitkovich2023rt,
  author    = {Zitkovich, Brianna and Yu, Tianhe and Xu, Sichun and Xu, Peng and Xiao, Ted and Xia, Fei and Wu, Jialin and Wohlhart, Paul and Welker, Stefan and Wahid, Ayzaan and others},
  title     = {{RT-2}: Vision-language-action models transfer web knowledge to robotic control},
  booktitle = {Proc. Conf. Robot Learn. (CoRL)},
  volume    = {229},
  pages     = {2165--2183},
  month     = {Nov},
  year      = {2023},
}

@inproceedings{kannan2024smart,
  title={{SMART-LLM}: Smart multi-agent robot task planning using large language models},
  author={Kannan, Shyam Sundar and Venkatesh, Vishnunandan LN and Min, Byung-Cheol},
  booktitle={Proc. IEEE/RSJ Int. Conf. Intell. Robots Syst. (IROS)},
  pages={12140--12147},
  year={2024},
}

@article{9547279,
  author  = {Abualigah, Laith and Diabat, Ali and Sumari, Putra and Gandomi, Amir H.},
  title   = {Applications, deployments, and integration of {Internet of Drones (IoD)}: A review},
  journal = {IEEE Sens. J.},
  year    = {2021},
  volume  = {21},
  number  = {22},
  pages   = {25532--25546},
  doi     = {10.1109/JSEN.2021.3114266}
}

@article{farquhar2024detecting,
  title={Detecting hallucinations in large language models using semantic entropy},
  author={Farquhar, Sebastian and Kossen, Jannik and Kuhn, Lorenz and Gal, Yarin},
  journal={Nature},
  volume={630},
  number={8017},
  pages={625--630},
  year={2024}
}

@inproceedings{NEURIPS2022_9d560961,
 author = {Wei, Jason and Wang, Xuezhi and Schuurmans, Dale and Bosma, Maarten and ichter, brian and Xia, Fei and Chi, Ed and Le, Quoc V and Zhou, Denny},
 booktitle = {Proc. Adv. Neural Inf. Process. Syst. (NeurIPS)},
 pages = {24824--24837},
 title = {Chain-of-Thought Prompting Elicits Reasoning in Large Language Models},
 volume = {35},
 year = {2022},
 month = {Dec},
}

@inproceedings{NEURIPS2023_271db992,
  author    = {Yao, Shunyu and Yu, Dian and Zhao, Jeffrey and Shafran, Izhak and Griffiths, Tom and Cao, Yuan and Narasimhan, Karthik},
  title     = {{Tree of Thoughts}: Deliberate problem solving with large language models},
  booktitle = {Proc. Adv. Neural Inf. Process. Syst.},
  volume    = {36},
  pages     = {11809--11822},
  year      = {2023},
  month     = {Dec}
}

@article{10591707,
  author  = {He, Ying and Fang, Jingcheng and Yu, F. Richard and Leung, Victor C.},
  title   = {Large language models ({LLMs}) inference offloading and resource allocation in cloud-edge computing: An active inference approach},
  journal = {IEEE Trans. Mobile Comput.},
  year    = {2024},
  volume  = {23},
  number  = {12},
  pages   = {11253--11264},
  doi     = {10.1109/TMC.2024.3415661}
}

@article{10818760,
  author  = {Zhang, Mingjin and Shen, Xiaoming and Cao, Jiannong and Cui, Zeyang and Jiang, Shan},
  title   = {{EdgeShard}: Efficient {LLM} inference via collaborative edge computing},
  journal = {IEEE Internet Things J.},
  year    = {2025},
  volume  = {12},
  number  = {10},
  pages   = {13119--13131},
  doi     = {10.1109/JIOT.2024.3524255}
}

@inproceedings{chen2024scalable,
  title={{Scalable Multi-Robot Collaboration with Large Language Models: Centralized or Decentralized Systems?}},
  author={Chen, Yongchao and Arkin, Jacob and Zhang, Yang and Roy, Nicholas and Fan, Chuchu},
  booktitle={Proc. IEEE Int. Conf. Robot. Autom. (ICRA)},
  pages={4311--4317},
  month = {May},
  year={2024},
}

@article{yang2025autohma,
  title={{AutoHMA-LLM}: Efficient task coordination and execution in heterogeneous multi-agent systems using hybrid large language models},
  author={Yang, Tingting and Feng, Ping and Guo, Qixin and Zhang, Jindi and Zhang, Xiufeng and Ning, Jiahong and Wang, Xinghan and Mao, Zhongyang},
  journal={IEEE Trans. Cogn. Commun. Netw.},
  volume={11},
  number={2},
  pages={987--998},
  year={2025},
}

@article{feng2025ochmas,
  title={{OC-HMAS}: Dynamic self-organization and self-correction in heterogeneous multiagent systems using multimodal large models},
  author={Feng, Ping and Yang, Tingting and Liang, Mingyang and Wang, Lin and Gao, Yuan},
  journal={IEEE Internet Things J.},
  volume={12},
  number={10},
  pages={13538--13555},
  year={2025},
}

\vfill

\end{document}